\documentclass[letterpaper, 10 pt, conference]{ieeeconf}  % Comment this line out if you need a4paper

\IEEEoverridecommandlockouts                              % This command is only needed if 
\usepackage{amsmath} % assumes amsmath package installed
\usepackage{amssymb}  % assumes amsmath package installed
\usepackage{todonotes}
\let\labelindent\relax % deactivate ieee labelindent that conflicts with enumitem
\usepackage{enumitem}
\usepackage{mathtools}
\usepackage{rotating}
\usepackage{multirow}
\usepackage{siunitx}
\usepackage{caption}
\usepackage{subcaption}
\usepackage{float}
\usepackage{hyperref}
\usepackage{pifont}
\usepackage{cite}
\usepackage{microtype}

\usepackage{booktabs} %nice table design

\usepackage{algorithmic} %to display algorithms
\usepackage{algorithm2e}
\graphicspath{{graphics/}}

\usepackage{tabto} % to use tabstops in lists

\hypersetup{
    colorlinks=true,
    linkcolor=black,
    filecolor=magenta,      
    urlcolor=cyan,
    pdftitle={Criticality Metric for VRUs},
    pdfpagemode=None,
    }

\title{\LARGE \bf
Scenario-Independent Criticality Assessment and Prediction for Vulnerable Road Users in Autonomous Driving
}

\author{Jörg Gamerdinger, Victor Schwarzenberger, Philipp Schmid, Sven Teufel and Oliver Bringmann
\thanks{University of T\"ubingen, Faculty of Science, Department of Computer Science, Embedded Systems Group 
\tt\small {\{joerg.gamerdinger, victor.schwarzenberger, philipp.schmid, sven.teufel, oliver.bringmann\} @uni-tuebingen.de}}
}%

\begin{document}
\maketitle
\thispagestyle{empty}
\pagestyle{empty}

\begin{abstract}

Increasing safety is the primary objective of automated vehicles. Achieving this goal requires reliable safety metrics that incorporate safety-relevant factors such as object type, velocity, and criticality. A key capability of such metrics is the distinction between critical and non-critical objects, which is addressed through criticality or relevance estimation. Existing criticality metrics are typically designed for specific scenarios and primarily focus on vehicle-to-vehicle interactions. In this paper, we therefore propose a novel criticality metric tailored to vulnerable road users (VRUs), which require special consideration due to their less predictable motion behavior. Furthermore, to avoid the complexity introduced by scenario-specific metrics, we introduce a scenario-independent criticality prediction framework applicable to all traffic participant classes. The effectiveness of both the proposed VRU-centric criticality metric and the criticality prediction framework is evaluated using the DeepAccident dataset, which contains a diverse set of safety-critical traffic scenarios.
The proposed VRU-centric criticality metric improves pedestrian criticality classification performance by up to 50\,\%. In addition, the proposed criticality prediction framework outperforms state-of-the-art metrics by 275\,\%, achieving an F1-score of 0.96 and enabling scenario-independent criticality assessment across all object classes. These results demonstrate the strong potential of the proposed approaches to enhance criticality assessment for safety evaluation in automated driving systems.

\end{abstract}

%%%%%%%%%%%%%%%%%%%%%%%%%%%%%%%%%%%%%%%%%%%%%%%%%%%%%%%%%%%%%%%%%%%%%%%%%%%%%%%%
\section{INTRODUCTION}
\label{sec:intro}

\begin{figure}[t]
    \centering
    \includegraphics[width=.9\linewidth]{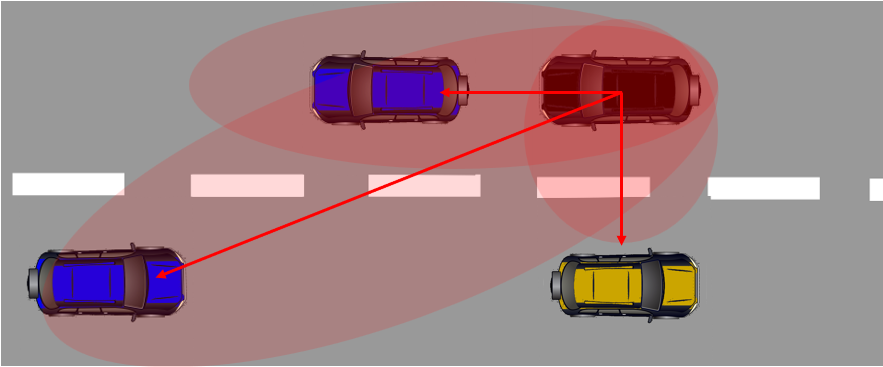}
    \caption{Exemplary Scene to represent criticality. The blue vehicles can be considered as critical or relevant as they could potentially collide with the ego (black). For the yellow vehicle a collision is unlikely; hence, the vehicle can be considered as not critical. Figure from~\cite{gamerdinger2025criticality}}
    \label{fig:criticality_example}
    \vspace*{-3mm}
\end{figure}

Road traffic accidents are the most likely cause of death for young people aged 5 to 29~\cite{Death}. Automated vehicles (AVs) are a promising approach to reducing accident rates, as human error is the main cause of fatal road accidents~\cite{EuropeanUnion2019}. In order to achieve safe automated driving, a complete and correct perception of the environment is crucial.  
Developing methods for object detection requires metrics to evaluate those methods and demonstrate their capabilities. However, state-of-the-art metrics such as average precision (AP) take into account all objects within a defined range and neglect safety-relevant information such as object types and velocities. Hence, novel metrics to evaluate the safety~\cite{volk2020safety, gamerdinger2025epsmnovelmetricevaluate} are proposed, which determine not only the performance, but the safety of object detection methods. One main difference between safety and performance evaluation is the inclusion of the criticality, which describes the necessity to perceive a specific object in order to avoid a safety-critical situation. 
As an example: A pedestrian crossing the street in front of the vehicle must be detected, as otherwise it could lead to a collision, while a vehicle driving behind the ego vehicle in an opposite direction could not directly lead to a safety-critical situation and, hence, can be considered as not critical. An exemplary scene with criticality classification is shown in Fig.~\ref{fig:criticality_example}.

To determine criticality, various metrics exist, as shown in different reviews~\cite{mahmud2017application, westhofen2023criticality}. Additionally, Gamerdinger et al.~\cite{gamerdinger2025criticality} performed a comprehensive evaluation, demonstrating the capabilities of the metrics in safety-critical situations. Within this work, it was identified that these metrics are highly scenario-dependent and do not generalize well. This requires the use of various metrics, which increases the complexity and reduces explainability. Moreover, those metrics are mostly defined for vehicle-vehicle interaction and neglect the increased vulnerability and lower predictability in motion of vulnerable road users (VRUs), such as pedestrians and bicyclists.

To overcome these limitations, in this work, we present a novel criticality metric focusing on the uncertainty in VRU movements. In addition, we propose a prediction framework which allows to assess criticality for all traffic participants, without the limitation to a specific scenario.

The main contributions of this work are:
\begin{itemize}
    \item We introduce the first dataset with object-dynamics-aware labels for perception criticality
    \item We propose a novel metric for criticality assessment of vulnerable road users
    \item We present a criticality prediction framework for a scenario-independent criticality assessment
\end{itemize}

In Sec.~\ref{sec:metrics_review}, we give an overview on current criticality metrics applicable to the safety assessment for object detection systems in automated driving. Section~\ref{sec:vru_method} presents our novel metric for VRUs, and Section~\ref{sec:prediction_method} a prediction method allowing for a scenario-independent criticality rating. A comprehensive evaluation, including a discussion of the results, can be found in Sec.~\ref{sec:eval}. Finally, we conclude our work and give an outlook.

\section{RELATED WORK}
\label{sec:metrics_review}

Criticality metrics for perception relevance can be divided into three categories: Time-based, Distance-based and Reachability-based. 

\paragraph{\textbf{Time-Based Metrics}}

The time-based criticality metrics determine the time to a specified event.
The most used metric to determine the criticality is the Time-to-Collision metric (TTC) as introduced by Hayward~\cite{hayward1972near} in 1972. Based on the distance and velocity, the time until a collision occurs is determined. The exact calculation depends on the relative positioning of the vehicles. For some situations TTC indicates $\infty$ as a result; however, the situation or object can be critical. Based on the TTC, the Crash Index Function (CIF) by Chan~\cite{chan2006defining} includes the severity of a potential collision in order to define the criticality. An improved version called modified TTC (MTTC) and Crash Index (CI) was presented by Ozbay et al.~\cite{ozbay2008derivation}. MTTC improves the calculation and the CI tries to estimate the criticality by the severity of the collision by including the relative velocities and accelerations. The Time-to-Accident (TTA) metric by Hydén~\cite{hyden1987development} calculates the time between an object starting an evasive maneuver and the collision. Similarly, the Time-to-Brake metric (TTB)~\cite{vanBrummelen2018} determines the time until a braking maneuver must be started to avoid a collision.
However, all these metrics only consider colliding vehicles as safety-critical and require a threshold to classify if an object is critical or not which makes them only partially suitable for criticality classification. 

\paragraph{\textbf{Distance-based Metrics}}
The most basic distance-based metric is the braking distance as used by Gamerdinger et al.~\cite{gamerdinger2024lsm}. They included a safety margin and the system delay to obtain a reliable safety distance that can be used to classify the criticality. However, the metric is used for lane detection and probably not optimal for dynamic object detection.
A widely used approach is the ``Responsible-Sensitive Safety'' (RSS) model presented by Shalev-Shwartz et al.~\cite{RSS}. The model aims to formalize human driving judgment in a mathematical model. Besides various safety distances for longitudinal and lateral directions and different directions of driving, the model consists of times and procedural rules which must be taken into account for an autonomous vehicle. Despite the different metrics for different scenarios, the metric does not achieve good results in classification and shows a high complexity in calculation~\cite{gamerdinger2025criticality}.
The SACRED metrics as proposed by Mori et al.~\cite{mori2023conservative} are a set of submetrics designed for motorway scenarios. The single submetrics are designed considering possible constellations of relative positions (same and opposite direction) and velocities (equal or different) on motorways. SACRED achieves high results for the classification of safety-critical objects; however, as it is designed based on motorway scenarios its application is limited.
An extension of the SACRED metrics was presented by Storms et al.~\cite{storms2023sure} with the SURE-Val metrics for urban scenarios. SURE-Val proposes three submetrics considering urban driving with static obstacles on the road and tangential cases such as a four-way intersection. In combination with SACRED, the metric set covers a wide range of scenarios and achieved good results as shown in~\cite{gamerdinger2025criticality}. However, the complexity and the limitation to very specific scenarios limits the applicability. In contrast to the time-based metrics, the distance-based metrics have the advantage that they directly output a threshold which allows to classify if an object is critical or not. However, these metrics show a higher complexity in calculation and mostly are designed for a specific scenario which also increases complexity in application.

\paragraph{\textbf{Reachability-based Metrics}}
A special case of simple distances are risk zones, which are defined by the possible vehicle trajectories. An object is then considered as critical if it is within this zone.
Canas et al.~\cite{canas2023dynamic} defined dynamic awareness criticality zones for parking maneuvers that are based on potential steering angles and distances to the vehicle. However, their method only determines a zone behind the vehicle for reverse parking scenarios up to \SI{5}{\kilo\meter\per\hour}.
A more general, reachability-based criticality zone model was presented by Topan et al.~\cite{topan2022interaction}. The model is capable of determining criticality zones based on potential control commands, vehicle kinematics up to \SI{20}{\metre\per\second}. However, they achieved not optimal results, and the calculation requires an offline precalculation of over \SI{200}{\second}, which makes it unsuitable for real application.

\section{PERCEPTION CRITICALITY DATASET}
\label{sec:dataset}
For the evaluation of criticality metrics, a dataset including safety-critical situations is required. Real-world accident datasets are not suitable, as they only contain information about the time of day, driver age, and so forth, as their principal objective is the prediction of accident severity. For criticality evaluation, relevant information, such as vehicle dimensions, velocities, and object classes, are not part of the datasets. Hence, we propose a criticality prediction dataset based on the synthetic DeepAccident dataset~\cite{Wang_2023_DeepAccident}. DeepAccident features 57K frames, divided into 690 scenarios. For each scenario, four vehicles equipped with sensors (ego, ego vehicle behind, other, other vehicle behind) are included. Based on NHTSA crash reports, the ego and the other vehicles collide in the accident scenarios. Besides those four vehicles, up to 46 other traffic participants of classes pedestrian, cyclist, motorcyclist, or vehicle are included as passive traffic ($\mu=13.82, \sigma=6.77$). As we require the provided meta data files, we use the 480 scenarios from the train split, which then leads to a total of 1920 scenarios, as the four sensor vehicles are taken separately. This leads to a total of 161,744 frames and 5,461,327 labeled objects. For each object, the object class, relative 3D position, dimension (length, width, height), yaw angle, relative velocity (x and y), the visibility from the camera and the visibility for the LiDAR, represented by the number of LiDAR points within the bounding box, is provided. In addition, two criticality labels are provided.
Currently, no standardized labeling for criticality classification exists. The main reason for this is the lack of a standardized definition. Hence, we use two distinct methods to determine the global criticality label: Multi-Metric-Based and Trajectory-Based.

\begin{itemize}
    \item \textbf{Multi-Metric-Based:} State-of-the-art metrics for criticality assessment are assumed to provide a correct classification. This fact is used within this work to create a safety-critical label. We evaluate different time and distance-based metrics. TTA and SACRED metrics showed a criticality indication of nearly \SI{100}{\percent}, which cannot be considered as reliable. Hence, we use the combination of $TTC_{\SI{3}{\second}}$, $CIF_{100}$ and $RSS_{0.5}$ as shown in Eq.~\eqref{eq:weak_gt} to determine the global criticality $C_o$ for an object $o$. 
    \begin{equation}
    \label{eq:weak_gt}
    C_{\mathrm{o}} = 
    \big(
    \mathrm{RSS}_{0.5}(o)
    \wedge
    \mathrm{TTC_{\SI{3}{\second}}} (o)
    \wedge
    \mathrm{CIF}_{100}(o) \big)
\end{equation}
where $TTC_{\SI{3}{\second}}$, $CIF_{100}$ and $RSS_{0.5}$ represent the binary result of the corresponding metric for the object $o$.
    
    For a detailed explanation on the metrics and the threshold, we refer to the comprehensive evaluation of Gamerdinger et al.~\cite{gamerdinger2025criticality}. As the metrics are trusted and well known, they mostly will provide a correct labeling; however, there remains some noise within these labels.
    
    \item \textbf{Trajectory-Based:} An object that collides with the ego can, without any doubt, be considered as safety-critical. However, not only colliding objects are critical, but also objects that could potentially collide or at least cross the ego vehicle's trajectory. For the ego vehicle, the trajectory for a limited time horizon (for the given scenario) can be considered as known. For all other objects, a reachability analysis is performed. For this, the current position, orientation, and a constant velocity model is assumed. For pedestrians, a circular-approximating polygon is determined, as a pedestrian could instantly change their direction, but a change in direction would lead to a lower distance, as it takes time to decelerate, turn around, and accelerate again. While this is not the most likely option, it must be considered from a safety perspective as a more conservative rating is to be preferred. The circular zone has the radius $r_{\text{pedzone}}$, as shown in Eq.~\eqref{eq:ped_zone}.
    \begin{equation}
        \label{eq:ped_zone}
        r_{\text{pedzone}} = t_{\text{traj}} \cdot v_{\text{ped}}
    \end{equation}
    $t_{\text{traj}}$ represents the time horizon of the ego trajectory, and $v_{\text{ped}}$ represents the velocity of the pedestrian.
    For all other traffic participants, a polygon is derived, since vehicles as well as bicyclists are not able to instantly change their direction of movement by any degree. The size of the polygon depends on an object type specific maximum steering angle, the velocity of the object, and $t_{\text{traj}}$. Regarding the steering angles for bicycles and motorcycles, $ \delta \pm \SI{22.5}{\degree}$ applies, while for cars, vans, and trucks, a speed-based maximum steering angle, $ \delta_{max,deg}$, is used, as defined in Eq.~\eqref{eq:steering_angle} by Rajamani~\cite{rajamani2006vehicle}. 

    \begin{equation}
        \label{eq:steering_angle}
        \begin{split}
        &R=\frac{v^2}{max(1e-6, \mu)\cdot 9.81}\\
        &\delta_{max,rad} = atan(R, D_{axle})\\
        &\delta_{max,deg} = \frac{\delta_{max,rad}\cdot 180}{\pi}\\
        \end{split}
    \end{equation}
    $\mu$ represents the road friction and $v$ the velocity of the vehicle. $D_{axle}$ represents the distance between the front and rear axle. The steering angle is limited to \SI{35}{\degree} for low velocities ($\leq \SI{15}{\kilo\meter\per\hour}$).
\end{itemize}

\section{VRU-CENTRIC CRITICALITY ASSESSMENT}
\label{sec:vru_method}
The metrics presented in Sec.~\ref{sec:metrics_review} are primarily designed for vehicle-to-vehicle interactions. However, vulnerable road users have a higher uncertainty in prediction due to their lower constraints to physics, such as friction coefficients or maximal accelerations. Especially, the lateral acceleration and possible turn rates are highly limited for vehicles, increasing the predictability, which makes state-of-the-art metrics suitable.

For the VRU-centric criticality assessment, a combination of behavior models for predicting pedestrian movement and an adaptation of the space occupancy index~\cite{johnsson2018search} is used. Pedestrians are able to abruptly stop and change their direction; this is a significant contrast to vehicles. This uncertainty must be included for the criticality assessment of pedestrians. For this purpose, we use the observations from Herman et al.~\cite{herman2021pedestrian}. They modeled the pedestrian movement with a circular shape that is slightly shifted towards the front, as changing the direction is less likely, and as well, the possible distance covered within the prediction horizon would be lower, since turning around and accelerating again takes some time. Considering the velocity of the pedestrians for the prediction, we use \SI{2.8}{\meter\per\second}, which corresponds to a fast walking. A slight overestimation of the velocity should be preferred in the context of safety. Using the position, the velocity, the orientation, and the prediction horizon, a polygon with the possible positions during the prediction horizon is defined. Then, the space occupancy is considered. For this purpose, also the possible space of the vehicle is derived, as described for the trajectory-based approach in Sec.~\ref{sec:dataset}.

\begin{figure*}[t!]
    \centering
    \begin{subfigure}[t]{0.33\textwidth}
        \centering
        \includegraphics[height=4.5cm]{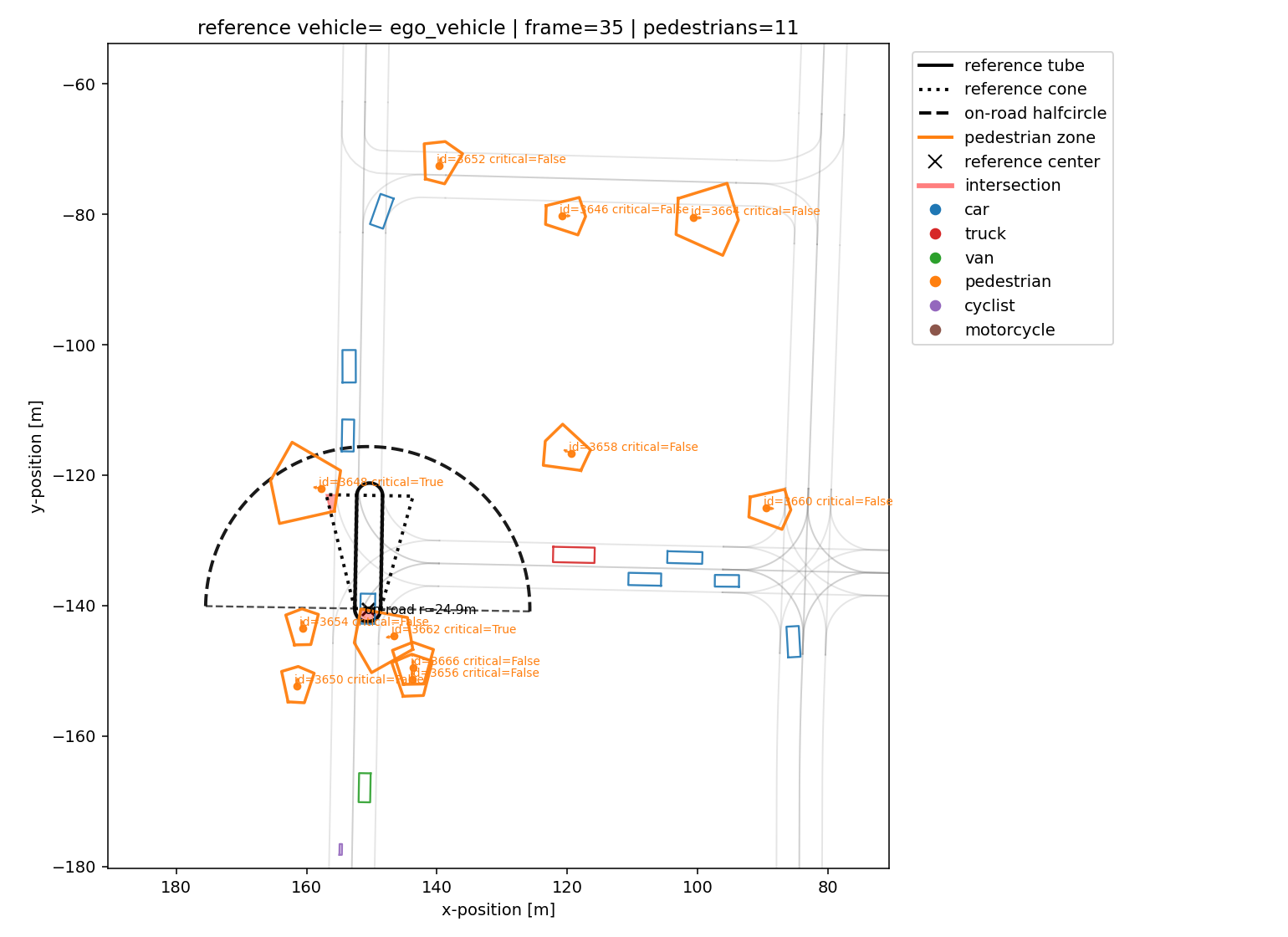}
        \caption{Scene 1}
    \end{subfigure}%
    \begin{subfigure}[t]{0.33\textwidth}
        \centering
        \includegraphics[height=4.5cm]{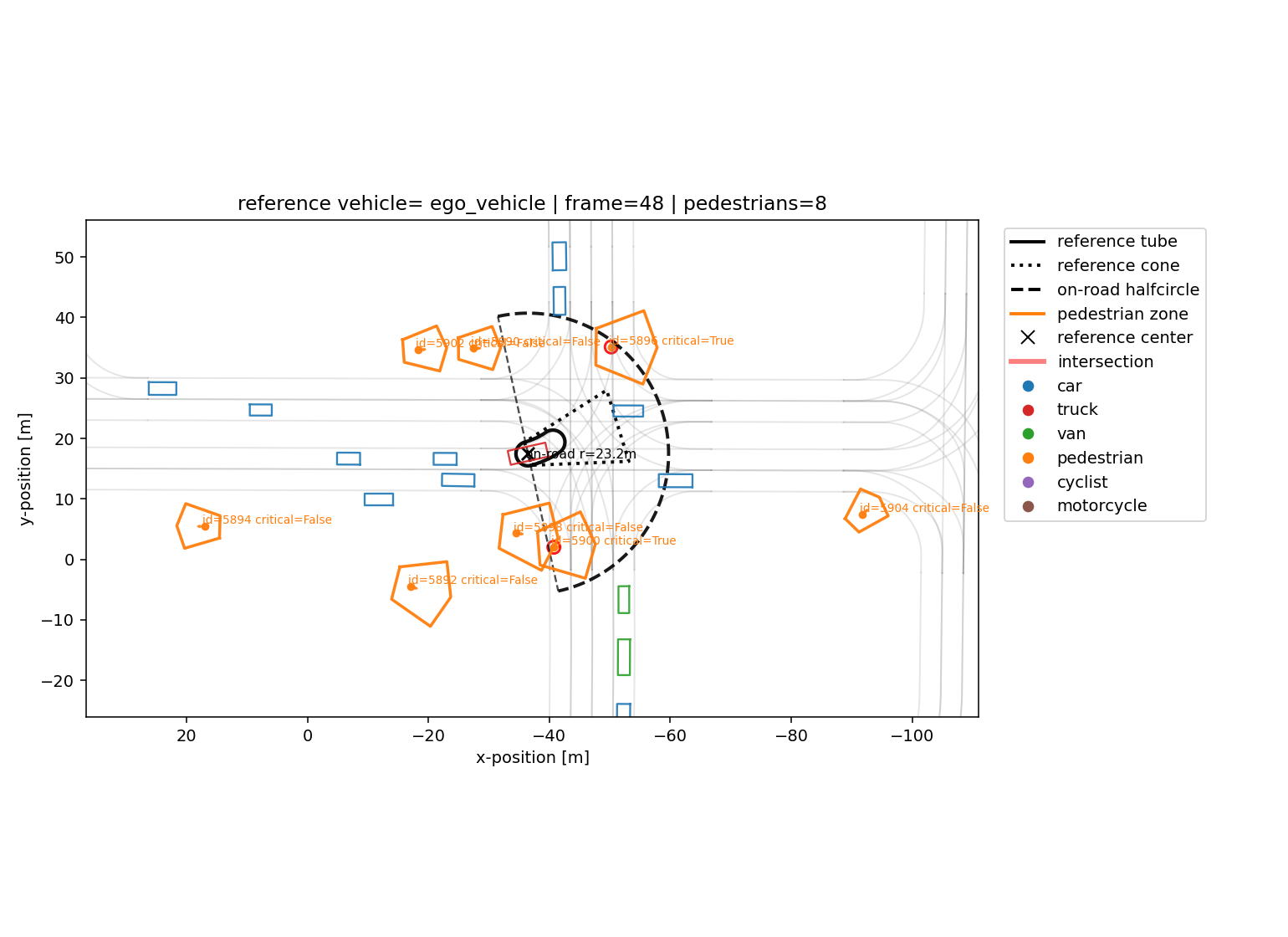}
        \caption{Scene 2}
    \end{subfigure}%
    \begin{subfigure}[t]{0.33\textwidth}
        \centering
        \includegraphics[height=4.5cm]{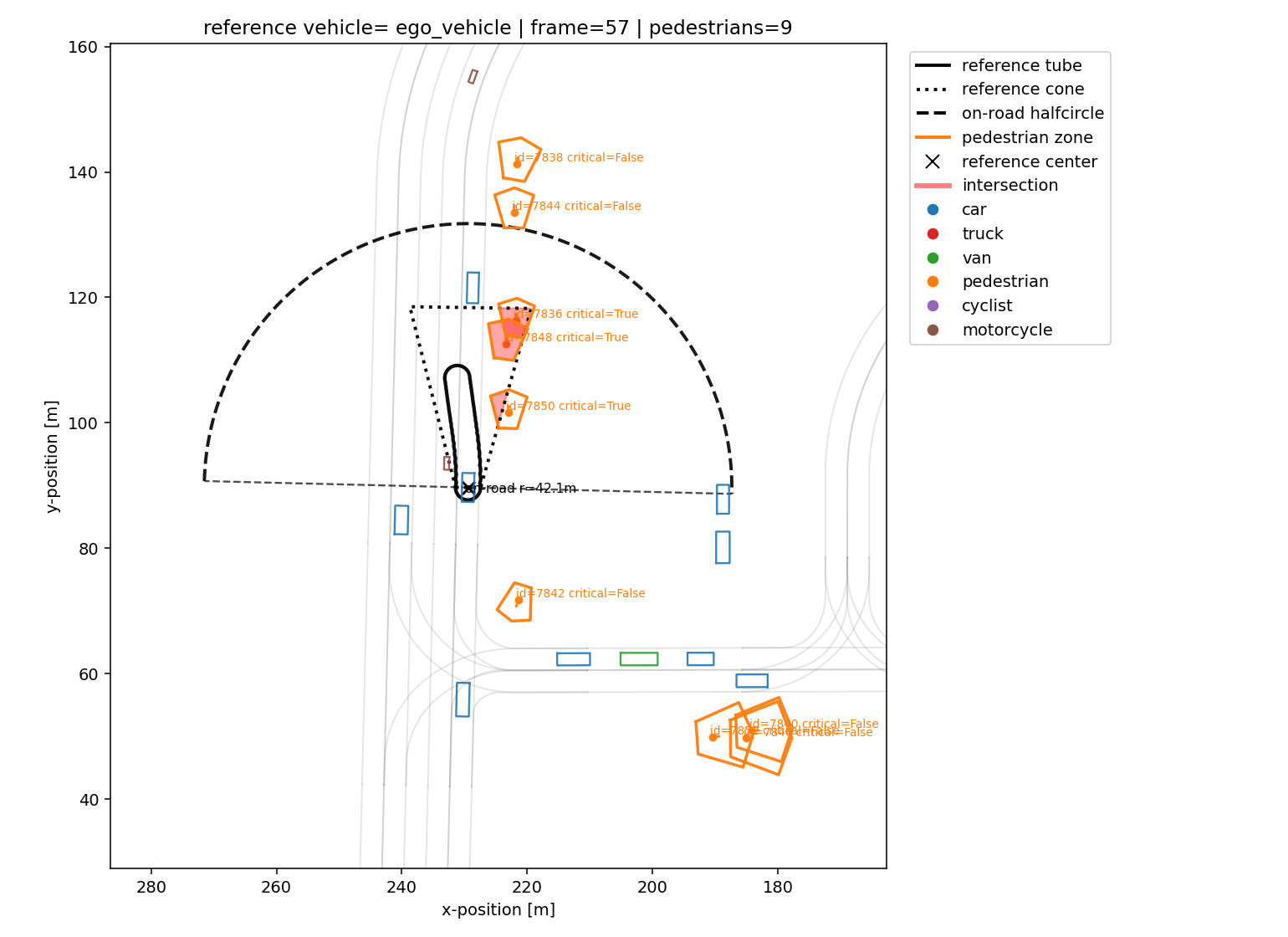}
        \caption{Scene 3}
    \end{subfigure}%

    \caption{Example scenes demonstrating the VRU criticality assessment}
    \label{fig:vru_metric}
    \vspace*{-5mm}
\end{figure*}

For an object $O_i$, the possible space at time $t$ is denoted as $SP(O_i,t)$. The space occupancy to determine the criticality is determined as described in Eq.~\eqref{eq:space}.

\begin{equation}
    \label{eq:space}
    Sp(A_i, t) \cap Sp(A_j, t) \neq \emptyset
\end{equation}

If this applies, no overlap is present and the object is considered uncritical. Otherwise, the VRU $A_j$ is considered critical.

In addition, pedestrians on the road must be always considered as critical within a suitable radius. Hence, pedestrians on the road within a distance of twice the required braking distance, depending on the given velocity of the ego vehicle, in driving direction are classified as critical. To avoid that there is no criticality zone for a vehicle while standing still, a minimum polygon size of \SI{6}{\meter} is defined which would include a pedestrian crossing when a vehicle has to stop before. 

This approach of combining the movement behavior of pedestrians with the space occupancy takes into account the specific properties in movement by pedestrians that are not considered in state-of-the-art criticality metrics. Exemplary scenes demonstrating the VRU criticality assessment are shown in Fig.~\ref{fig:vru_metric}.

\section{CRITICALITY PREDICTION} % potentiell was cooles ergänzen
\label{sec:prediction_method}
The evaluation in~\cite{gamerdinger2025criticality} has shown, that different metrics are not able to generalize across different scenarios. Machine learning is a promising approach to overcome these limitations and provide a better generalization over road users and scenarios if a suitable dataset exists. Using the dataset presented in Sec.~\ref{sec:dataset}, we are able to propose an object-class and scenario-independent prediction framework using statistical models.
For the prediction four unsupervised and four supervised methods are employed. The unsupervised methods are KMeans~\cite{macqueen1967kmeans}, DBScan~\cite{ester1996dbscan}, Isolation Forest (IF)~\cite{liu2008isolation} and Gaussian mixture model (GMM)~\cite{mclachlan2019finite}. The supervised methods are linear regression (LR)~\cite{montgomery2021introduction}, random forest (RF)~\cite{breiman2001random}, histogram-based gradient boosting (HGB)~\cite{friedman2001gradientboosting} and LightGBM~\cite{ke2017lightgbm}. 

This collection of models includes clustering approaches as well as regression and tree-based approaches for predicting the criticality. More complex models would reduce the traceability which is inappropriate in terms of safety. Hence, simpler statistical models are used.

As input features for the prediction the input vector $V$ as defined in Eq.~\eqref{eq:feature_vector} is used.
\begin{equation}
    \label{eq:feature_vector}
    V = (c,x,y,z,l,w,h,\Psi,v_x,v_y,lp,vis)
\end{equation}
$c$ represents the object class, $x,y,z$ describe the relative position of the object to the ego, $l,w,h$ the dimension, $\Psi$ the relative yaw angle, $v_x, v_y$ the relative velocity, $lp$ the number of LiDAR points in the bounding box and $vis$ represents a boolean describing if the object is visible from the camera perspective. 

All supervised models are trained on a layered 80/20 train-test split. As non-critical objects naturally dominate real traffic scenes, the class distribution reflects this imbalance; balanced class weights are applied during training to ensure that the minority class receives adequate attention. Input features are standardized for LR. For RF and LightGBM, hyperparameters are tuned using Optuna~\cite{optuna_2019} with a Tree-structured Parzen Estimator sampler, optimizing F1 via stratified 3-fold cross-validation on the training set to prevent test set leakage. The search covers parameters including the number of estimators, tree depth, learning rate, and regularization terms. Furthermore, for all supervised models, the decision threshold is tuned by scanning the range $[0.01, 0.99]$ and selecting the value maximizing F1 on the test set.

\section{RESULTS \& DISCUSSION}
\label{sec:eval}
\subsection{Experiments}
Both the VRU-centric criticality metric and the criticality prediction are evaluated using the dataset proposed in Sec.~\ref{sec:dataset}.
For evaluation, the precision $P$ and recall $R$, as shown in Eq.~\eqref{eq:prec}, and their harmonic mean $F1$, as defined in Eq.~\eqref{eq:f1}, are employed.

\begin{equation}
    \label{eq:prec}
    P = \frac{TP}{TP+FP},\quad R = \frac{TP}{TP+FN}
\end{equation}
\begin{equation}
    \label{eq:f1}
    F1 = 2\cdot\frac{P\cdot R}{P+R}
\end{equation}
$TP$ represents a correct classification as critical, $TN$ a correct classification as not critical, $FP$ an erroneous classification as critical, and $FN$ an erroneous classification as uncritical.

\begin{figure}[H]
    \centering
    \includegraphics[width=1\linewidth]{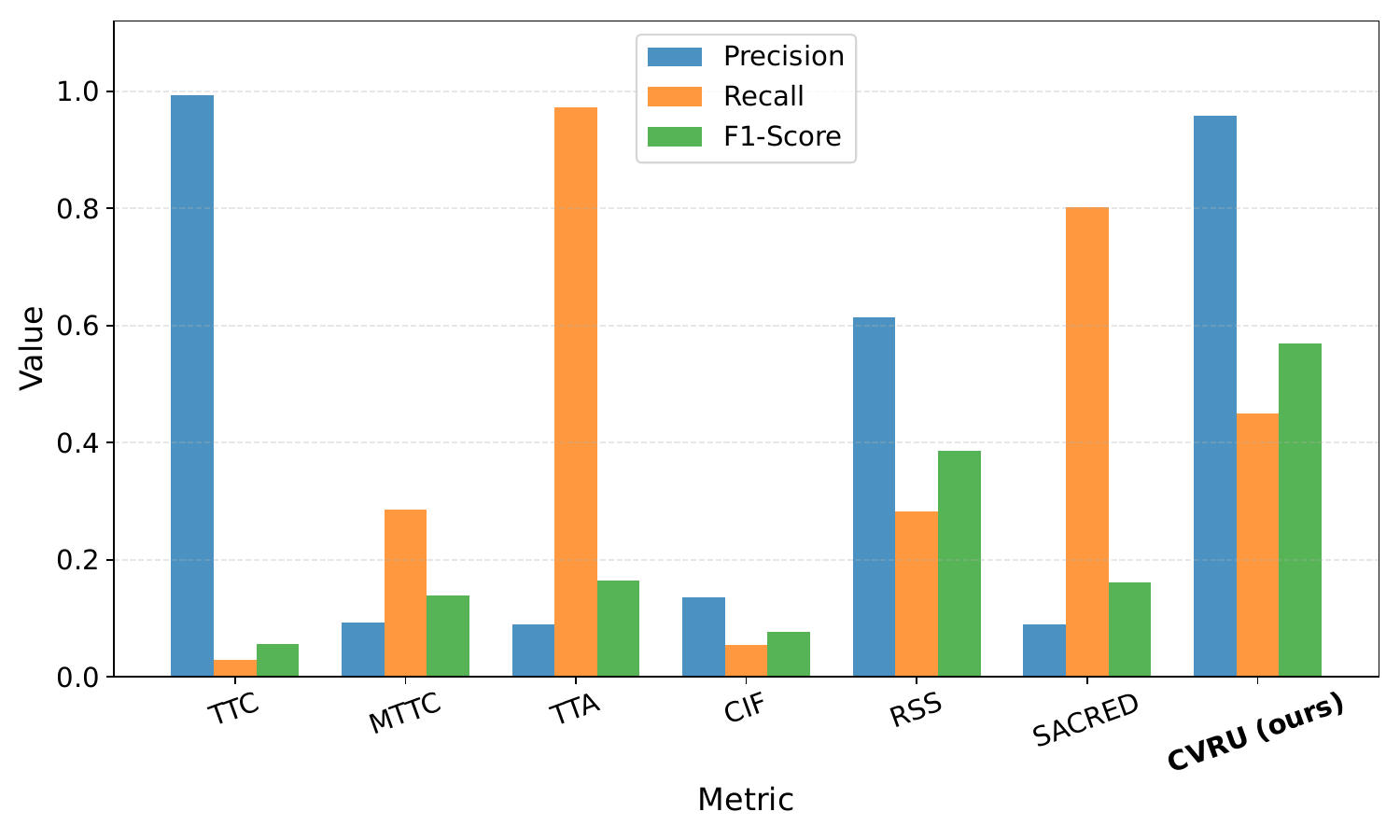}
    \caption{Results on pedestrian class for state-of-the-art metrics and our novel VRU-centric metric CVRU.}
    \label{fig:vru_plot}
\end{figure}

\begin{figure}
    \centering
    \includegraphics[width=1\linewidth]{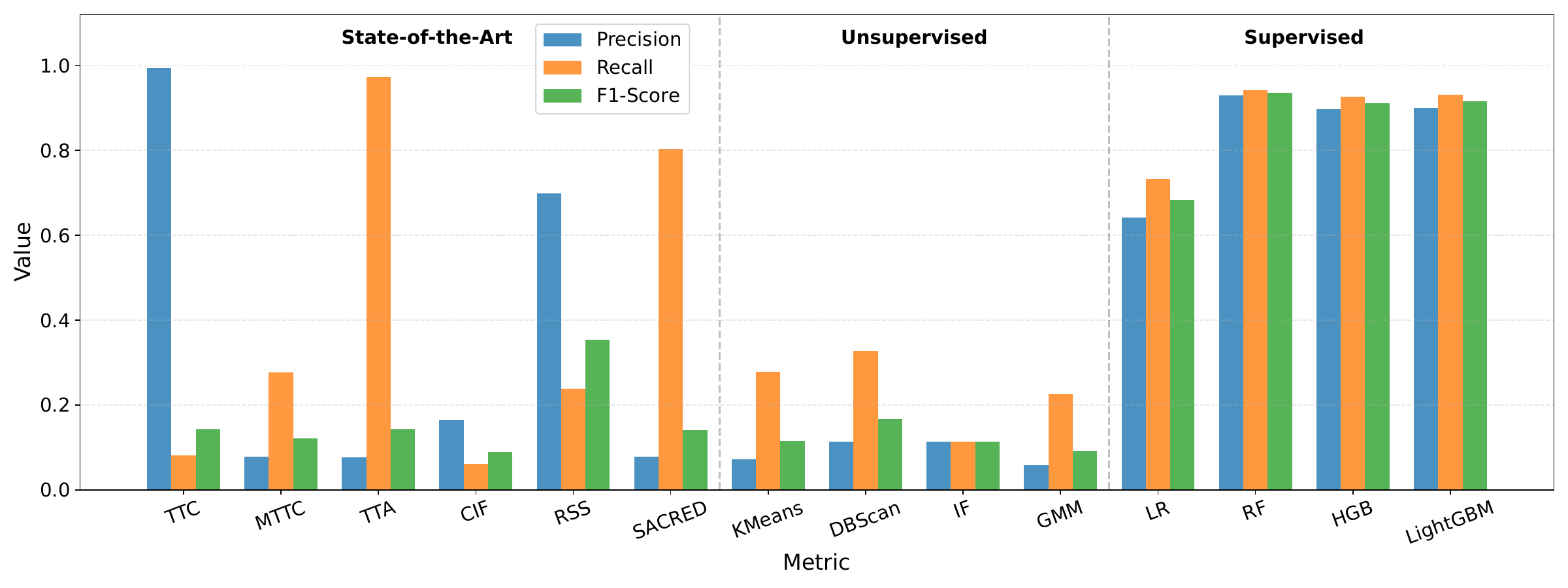}
    \caption{Results of the state-of-the-art metrics compared to the prediction methods.}
    \label{fig:prediction_plot}
\end{figure}

Results are visualized in Fig.~\ref{fig:vru_plot} and Fig.~\ref{fig:prediction_plot} and summarized in Tab.~\ref{tab:results_vru} and Tab.~\ref{tab:results_prediction}. For the criticality prediction, the methods introduced in Sec.~\ref{sec:prediction_method} are employed. As comparison metrics, TTC, MTTC, TTA, CIF, RSS, and SACRED, as introduced in Sec.~\ref{sec:metrics_review}, were chosen as they performed best in our previous work~\cite{gamerdinger2025criticality}.

\subsection{VRU-Centric Criticality Metric}
The state-of-the-art (SOTA) methods are, as discussed, designed for vehicle-vehicle interaction and neglect VRU-specific properties. This can also be observed in the classification results. The TTC achieved a high precision, with 0.99, however, a really low recall, leading to an F1-score of only about 0.06. CIF performed slightly better, with about 0.08. MTTC, TTA, and SACRED achieved F1-scores of about 0.15, which is a significant increase compared to TTC and CIF; however, can be considered as not suitable. RSS achieved the highest F1-score of all SOTA metrics with 0.39, and a precision and recall of 0.61 and 0.28, respectively. Our VRU-centric CVRU metric significantly outperformed the SOTA metrics. While the precision with 0.96 is 0.03 lower as for TTC, the recall with 0.45 and F1-score with 0.57 are the best results. This corresponds to an increase in F1-score of about \SI{50}{\percent} compared to RSS as best SOTA metric.
\begin{table}[t]
      \caption{Results on the proposed dataset. Results are Precision (P), Recall (R) and F1-score (F1) for the pedestrian class.}
      \centering
    \begin{tabular}{l    rrrr} \toprule
    Metric& P & R & F1 \\ \midrule
    TTC & \textbf{0.9931}& 0.0294& 0.0560\\
    MTTC &  0.0921& 0.2857& 0.1392\\
    TTA &  0.0896& 0.9730& 0.1642\\
    CIF & 0.1366& 0.0545& 0.0774\\
    RSS &  0.6135& 0.2821& 0.3865\\
    SACRED & 0.0895 & 0.8019& 0.1611\\
    \midrule
    CVRU (ours) & 0.9587& \textbf{0.4494}& \textbf{0.5687}\\
    \bottomrule
\end{tabular}
\vspace*{0mm}
\label{tab:results_vru}
\end{table}

\subsection{Criticality Prediction}

Considering all classes, the state-of-the-art (SOTA) methods achieved F1-scores ranging from 0.0883 for CIF to 0.3545 for RSS, demonstrating that they are not suitable for safety-critical classification. Similarly, the unsupervised methods achieved only limited performance, with F1-scores ranging from 0.0922 for GMM to 0.1678 for DBScan.

In contrast, the supervised methods showed a significant improvement. Logistic Regression achieved an F1-score of 0.6836, representing an improvement of more than 0.32 compared to the best-performing SOTA metric. The performance was further improved using Histogram-based Gradient Boosting (HGB), which achieved a precision of 0.8975, a recall of 0.9263, and an F1-score of 0.9117. LightGBM provided a slight improvement over HGB, achieving a precision of 0.9005, a recall of 0.9317, and an F1-score of 0.9158. The best overall performance was achieved by the Random Forest classifier, reaching a precision of 0.9303, a recall of 0.9415, and an F1-score of 0.9359. This corresponds to an improvement of approximately 0.58 in F1-score compared to the best SOTA metric. The superior performance of the supervised approaches can be attributed to their ability to learn complex relationships from data, resulting in substantially better generalization than handcrafted SOTA criticality metrics. The minor outperform in precision from TTC and in recall from TTA can be traced back to some oversensitivity of these metrics and the natural imbalance of the dataset.

\subsection{Discussion}
The obtained results demonstrate that state-of-the- metrics are not suitable for the criticality classification of pedestrians. By leveraging the proposed CVRU metric, which incorporates the specific characteristics of pedestrians, including their higher degree of freedom in movement, significantly higher classification performance can be achieved. Furthermore, the SOTA metrics exhibited considerable limitations when applied across diverse traffic scenarios, confirming the observations reported in~\cite{gamerdinger2025criticality}. Using a tree-based prediction model, an F1-score of up to 0.9359 was achieved, demonstrating that the proposed approach is well suited for pedestrian criticality classification.
%%%%%%%%%%%%%%%%%%%%%%%%%%%%%%%

\begin{table}
      \caption{Results on the proposed dataset. Results are Precision (P), Recall (R) and F1-score (F1) over all classes.}
      \centering
    \begin{tabular}{l    rrr} \toprule
    Metric& P & R & F1 \\ \midrule
    \multicolumn{4}{c}{\textbf{State-of-the-Art}}\\
    TTC &  0.9939& 0.0817& 0.1429\\
    MTTC & 0.0775& 0.2770& 0.1211\\
    TTA &  0.0772& 0.9722& 0.1430\\
    CIF &  0.1651& 0.0606& 0.0883\\
    RSS &  0.6985& 0.2375& 0.3545\\
    SACRED & 0.0775 & 0.8035& 0.1412\\
    \midrule
    \multicolumn{4}{c}{\textbf{Unsupervised}}\\
    KMeans & 0.0725 & 0.2787 & 0.1150\\
    DBScan & 0.1128 & 0.3279 & 0.1678\\
    IF &  0.1141 & 0.1141 & 0.1141\\
    GMM & 0.0579 & 0.2263 & 0.0922\\
    \midrule
    \multicolumn{4}{c}{\textbf{Supervised}}\\
    LR & 0.6411 & 0.7321 & 0.6836\\
    RF & \textbf{0.9303} & \textbf{0.9415} & \textbf{0.9359}\\
    HGB & 0.8975 & 0.9263 & 0.9117\\
    LightGBM & 0.9005 & 0.9317 & 0.9158\\
    \bottomrule
\end{tabular}
\vspace*{0mm}
\label{tab:results_prediction}
\end{table}

%%%%%%%%%%%%%%%%%%%%%%%%%%%%%%%%%%%%%%%%%%%%%%%%%

\section{CONCLUSION \& OUTLOOK}
\label{sec:conclusion}

In this paper, we introduced the first perception-criticality dataset together with a novel VRU-centric criticality metric and a scenario-independent criticality prediction framework. The proposed dataset is based on the DeepAccident dataset and comprises 161k frames with 5.46M labeled objects. For each frame and object, the outputs of various criticality metrics are provided. The criticality labels are generated using a metric-aggregation approach combined with geometric properties and kinematic constraints of traffic participants. The proposed VRU-centric criticality metric explicitly accounts for the uncertainty in VRU motion and their increased freedom of movement and direction changes. To this end, VRU motion models are combined with the Space Occupancy Index approach, where potential space intersections are used to estimate criticality. Using this VRU-specific metric, the classification performance in terms of F1-score improves by up to \SI{50}{\percent} compared to the best state-of-the-art metric. To address the limitations of object-class-dependent and scenario-specific metrics, we further proposed a scenario-independent criticality prediction framework based on supervised and unsupervised clustering, regression, and tree-based models. Evaluation on the proposed dataset demonstrates an F1-score improvement of 0.60, corresponding to an increase of approximately \SI{275}{\percent} in criticality classification performance.

Future work will focus on extending the prediction framework with additional prediction methods and further fine-tuning to improve performance. In addition, temporal information and the proposed VRU-centric metric will be incorporated to further enhance criticality prediction accuracy. The dataset and the criticality prediction framework will be publicly released to support future research and contribute to improving safety in autonomous driving. Furthermore, more detailed analyses for individual object classes will be provided.

%%%%%%%%%%%%%%%%%%%%%%%%%%%%%%%%%%%%%%%%%%%%%%%%%%%%%%%%%%%%%%%%%%%%%%%%%%%%%%%%

%%%%%%%%%%%%%%%%%%%%%%%%%%%%%%%%%%%%%%%%%%%%%%%%%%%%%%%%%%%%%%%%%%%%%%%%%%%%%%%%

%%%%%%%%%%%%%%%%%%%%%%%%%%%%%%%%%%%%%%%%%%%%%%%%%%%%%%%%%%%%%%%%%%%%%%%%%%%%%%%%
%\section*{APPENDIX}

%Appendixes should appear before the acknowledgment.

%%%%%%%%%%%%%%%%%%%%%%%%%%%%%%%%%%%%%%%%%%%%%%%%%%%%%%%%%%%%%%%%%%%%%%%%%%%%%%%%

\bibliographystyle{IEEEtran} % use IEEEtran.bst style
\bibliography{literature.bib}

\end{document}